\documentclass[runningheads]{llncs}

\usepackage{caption}
\usepackage{graphicx}
\usepackage{booktabs}
\usepackage{placeins}
\usepackage[accsupp]{axessibility} 
\usepackage{orcidlink}
\usepackage{algorithm}
\usepackage{algorithmicx}
\usepackage{algpseudocode}
\usepackage{tcolorbox}
\usepackage{booktabs}
\usepackage[table]{xcolor}
\usepackage{float}
\usepackage{hyperref}

\newcommand{\yd}[1]{{\textcolor{black}{#1}}}
\newcommand{\yang}[1]{{\textcolor{black}{#1}}}
\newcommand{\deng}[1]{{\textcolor{black}{#1}}}
\newcommand{\yl}[1]{{\textcolor{black}{#1}}}
\newcommand{\dy}[1]{{\textcolor{black}{#1}}}

\begin{document}

% \title{From Guidance to Selection: Training-Free Object-Background Compositional T2I 
% via Dynamic Spatial Guidance and Multi-Path Pruning} 

\title{Training-Free Object-Background Compositional T2I 
via Dynamic Spatial Guidance and Multi-Path Pruning} 
\titlerunning{Object-Background Compositional T2I}

\author{Yang Deng\inst{1}\orcidlink{0009-0007-9162-1759} \and
David Mould\inst{2}\orcidlink{0000-0001-5779-484X} \and
Paul L. Rosin\inst{2}\orcidlink{0000-0002-4965-3884} \and
Yu-Kun Lai\inst{1}\orcidlink{0000-0002-2094-5680}}

\authorrunning{Y. Deng et al.}

% \institute{Cardiff University, Cardiff, NJ 08544, USA \and
% Springer Heidelberg, Tiergartenstr.~17, 69121 Heidelberg, Germany
\institute{
Cardiff University, Cardiff, United Kingdom\\
\email{\{DengY9, RosinPL, LaiY4\}@cardiff.ac.uk}
\and
Carleton University, Ottawa, Canada\\
\email{mould@scs.carleton.ca}
}
\maketitle
\begin{abstract}
Existing text-to-image diffusion models, while excelling at subject synthesis, exhibit a persistent foreground bias that treats the background as a passive and under-optimized byproduct. This imbalance compromises global scene coherence and constrains compositional control.
To address the limitation, we propose a training-free framework that restructures diffusion sampling to explicitly account for foreground–background interactions. Our approach consists of two key components. First, \textit{Dynamic Spatial Guidance} introduces a soft, time step dependent gating mechanism that modulates foreground and background attention during the diffusion process, enabling spatially balanced generation. 
Second, \textit{Multi-Path Pruning} performs multi-path latent exploration and dynamically filters candidate trajectories using both internal attention statistics and external semantic alignment signals, retaining trajectories that better satisfy object–background constraints. We further develop a benchmark specifically designed to evaluate object-background compositionality.
Extensive evaluations across multiple diffusion backbones demonstrate consistent improvements in background coherence and object–background compositional alignment.

\keywords{Text to Image \and Diffusion Model \and Dynamic Pruning}
\end{abstract}
% \vspace{-4mm}
\section{Introduction}
\label{sec:intro}
\vspace{-4mm}
\yang{Diffusion-based text-to-image models have demonstrated remarkable progress in synthesizing multiple entities with high fidelity. However, recent advances have focused on \emph{object-centric} composition~\cite{wang2025not,huang2023t2i,zhang2024realcompo,qu2025silmm,li2022stylet2i}} with prompts such as “a man riding a wave on top of a surfboard”~\cite{hu2023tifa}, “a red book and a yellow vase”~\cite{hu2024ella}, or “a metallic car and a wooden desk”~\cite{huang2023t2i}, emphasizing the coexistence of \yang{multiple objects} and accurate attribute binding.
\yang{In these object-centric studies, the} foreground has been treated as the bearer of semantic content, whereas the \emph{background}

is often considered decorative noise or uncontrolled texture.

Not only is the background valuable in its own right,
coherent foreground-background relations are essential for compositional integrity~\cite{poore1976composition}. 
In this paper, we address the underexplored challenge of ensuring coherent interaction between a main subject and its complex background. In text-to-image \yl{(T2I)} generation, prompts typically describe both foreground and background, including the subject’s attributes (category, appearance, pose) and contextual factors such as layout, lighting, and surrounding objects. Our goal is to generate images where the subject and its environment are spatially and semantically coherent, and adhere to the input prompt.
This challenge is not only aesthetic: the background must semantically reflect the elements and scene context described in the prompt, providing cues that shape both meaning and visual realism. The importance of such context is evident in object recognition research, where models can achieve high accuracy using only background cues~\cite{oliva2007role,torralba2003contextual,barbu2019objectnet}, and adversarial backgrounds alone can cause misclassification rates of up to 87.5\%~\cite{xiao2020noise}.

Prior efforts address the compositional challenges involving multiple foreground objects by isolating each object and relying on large language models (LLMs) to propose \yang{explicit geometric layouts}, followed by region-wise generation and subsequent fusion~\cite{yang2024mastering,chen2024region}.
While conceptually appealing, these systems rely on externally imposed layouts~\cite{feng2023layoutgpt,zheng2023layoutdiffusion}, which often conflict with the spatial priors encoded in the initial noise. Such mismatches weaken the generative prior, leading to degraded visual fidelity or semantic drift~\cite{zhang2024realcompo}. Moreover, many pipelines require \yl{an} explicit bounding \yl{box} for each object in the prompt, resulting in a rigid design that depends on LLM-generated structures often prone to errors. To address these issues, Be Decisive~\cite{dahary2025decisive} employs a neural predictor to estimate object positions. However, it assumes that the initial noise latent provides reliable spatial priors, which is not always valid, and struggles with complex overlaps or occlusions.

\begin{figure}[t]
    \centering
    \includegraphics[width=\linewidth]{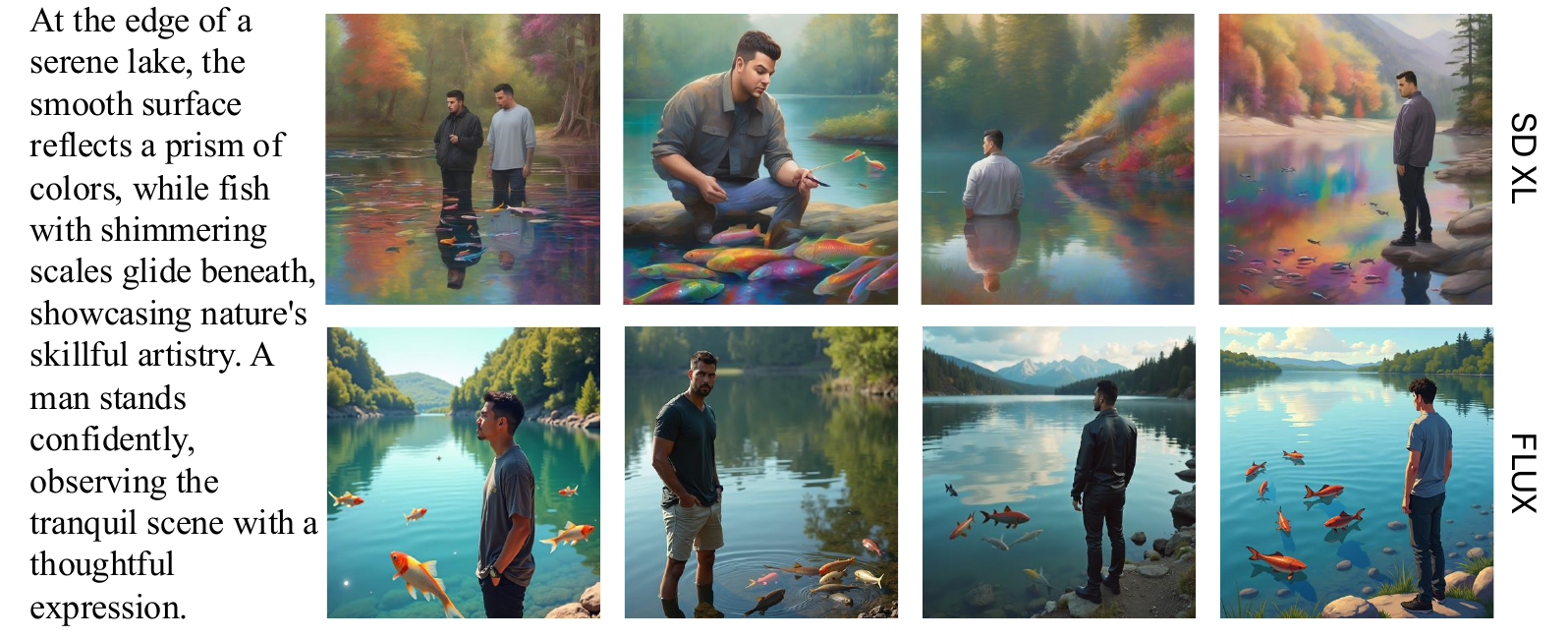}
    \vspace{-5mm}
    \caption{
    Seed sensitivity of Stable Diffusion XL~\cite{podell2023sdxl} and FLUX~\cite{flux2024}. With the same prompt, different random seeds produce distinct layouts and semantics. Each row shows four results from different seeds for one model.}
    \vspace{-6mm}
    \label{fig:differentseeds}
\end{figure}

As shown in Fig.~\ref{fig:differentseeds}, diffusion models exhibit strong seed-dependent instability. Different random seeds generate distinct initial noise fields whose low-frequency structures implicitly determine the global layout~\cite{ban2024crystal,guo2024initno,patashnik2023localizing}. Consequently, identical prompts can yield diverse compositional outcomes.

Inspired by the ``multi-path reasoning'' principle~\cite{fu2025deep}, \yang{in which robust inference emerges from generating multiple reasoning paths and selecting among them based on internal confidence, }we reconsider compositionality not as a single deterministic sampling, but as a dynamic competition among multiple plausible scene hypotheses.

% Building on this insight, 
In this work, we introduce a training-free, multi-path diffusion framework that reframes compositional generation as selection among competing noise trajectories.
Our method explicitly decouples the background and object formation phases: during early denoising, a Dynamic Spatial Guidance (DSG) module extracts background-aware attention masks to softly anchor global structure, while later steps refine object details under consistent spatial context.
Across multiple stochastic paths, a Dynamic Pruning strategy continuously evaluates semantic focus and prompt fidelity, preserving only coherent trajectories. 

\dy{In summary, our main contributions are: 1) We study the task of \textit{Object--Background Compositionality} in text-to-image generation, where foreground subjects and background scenes jointly follow complex prompts.
2) We propose a \textit{training-free framework} that improves object–background coherence during sampling, 
introducing \textit{Dynamic Spatial Guidance (DSG)} for attention-derived spatial decoupling and \textit{Multi-Path Dynamic Pruning} that selects reliable trajectories using internal attention statistics and vision–language alignment.
3) We establish a \textit{benchmark} for this task, including a structured prompt dataset and an evaluation protocol for compositional fidelity.
4) We demonstrate the generality of our framework across diverse diffusion backbones, consistently achieving state-of-the-art performance on the proposed benchmark.}

% \vspace{-4mm}

\section{Related Work}
% \vspace{-4mm}
Text-to-image (T2I) diffusion models deliver striking visual fidelity and prompt alignment~\cite{nichol2021glide,rassin2022dalle,dhariwal2021diffusion,podell2023sdxl,chen2023pixart,flux2024}, yet structurally coherent compositional generation remains elusive for prompts with multiple entities, attributes, and spatial relations. \yl{The} first stream pursues numeric layout control~\cite{chen2024region,yang2024mastering,li2025mccd}, where large language models \yl{(LLMs)} or planners predict coordinates such as boxes or masks to enforce spatial placement during denoising. \yang{Although these methods offer explicit spatial guidance, the externally fixed layouts can interfere with the model’s intrinsic spatial prior derived from noise initialization, leading to degraded realism or semantic drift~\cite{zhang2024realcompo}. In addition, their reliance on LLM-generated coordinates or masks makes them brittle, as structured responses are often noisy or ill-formed.}

To relax hard geometry, semantic layout methods~\cite{feng2024ranni,wang2024genartist,zhang2025layercraft} \yang{manipulate text embeddings} that encode objects, attributes, and relations. This improves semantic disentanglement and robustness but typically remains object-centric and lacks explicit spatial grounding, which limits handling of fine-grained geometry, occlusion, and the integration of ambience with subjects, where ambience refers to global scene factors such as lighting~\cite{ren2024relightful}, color harmony~\cite{hertz2024style}, and environmental context that determine aesthetic coherence~\cite{zhang2024realcompo}.

A complementary line of work targets faithfulness through \yl{improved encoders} or objective design rather than layout control. Stronger text encoders and feedback-driven optimization~\cite{qu2025silmm,liu2025llm4gen,meral2024conform} enhance image–text agreement, and fine-tuning strategies~\cite{bao2024separate,wang2025energy} aim to disentangle compositional subspaces. Despite gains, these approaches require additional supervision or retraining, which reduces generality and efficiency. Benchmarks~\cite{hu2024ella,huang2023t2i,hu2023tifa} further show a bias toward salient foreground entities, with contextual dependencies and background semantics under-modeled. 
Energy-based formulations~\cite{zhang2024object,wang2025energy} reinterpret cross-attention as minimizing learned scene potentials, offering structural interpretability and improved object–attribute alignment. However, the learned energy landscapes remain largely static at inference, and thus fail to capture the time-evolving spatial dependencies between foreground and background regions that emerge throughout the denoising trajectory.

Recent evidence~\cite{guo2024initno,ban2024crystal,patashnik2023localizing} highlights the role of the initial noise in shaping low-frequency structure and global layout early in denoising, indicating that composition is tightly linked to sampling dynamics. Be Decisive~\cite{dahary2025decisive} leverages this perspective by training a small predictor to estimate object positions from noise priors, which improves subject separation but assumes sufficiently informative initial latents and struggles with complex overlap or occlusion. Moreover, existing studies rarely model the interplay between foreground objects and the background environment, leaving open how to achieve composition that harmonizes subjects with ambience while preserving generative diversity. This gap motivates a training-free, sampling-level perspective that explicitly reasons over multiple trajectories and foreground-background coupling, setting the stage for our approach.
% \vspace{-6mm}
\section{Method}
% \vspace{-4mm}
\begin{figure*}[t]
    \centering
    \includegraphics[width=\linewidth]{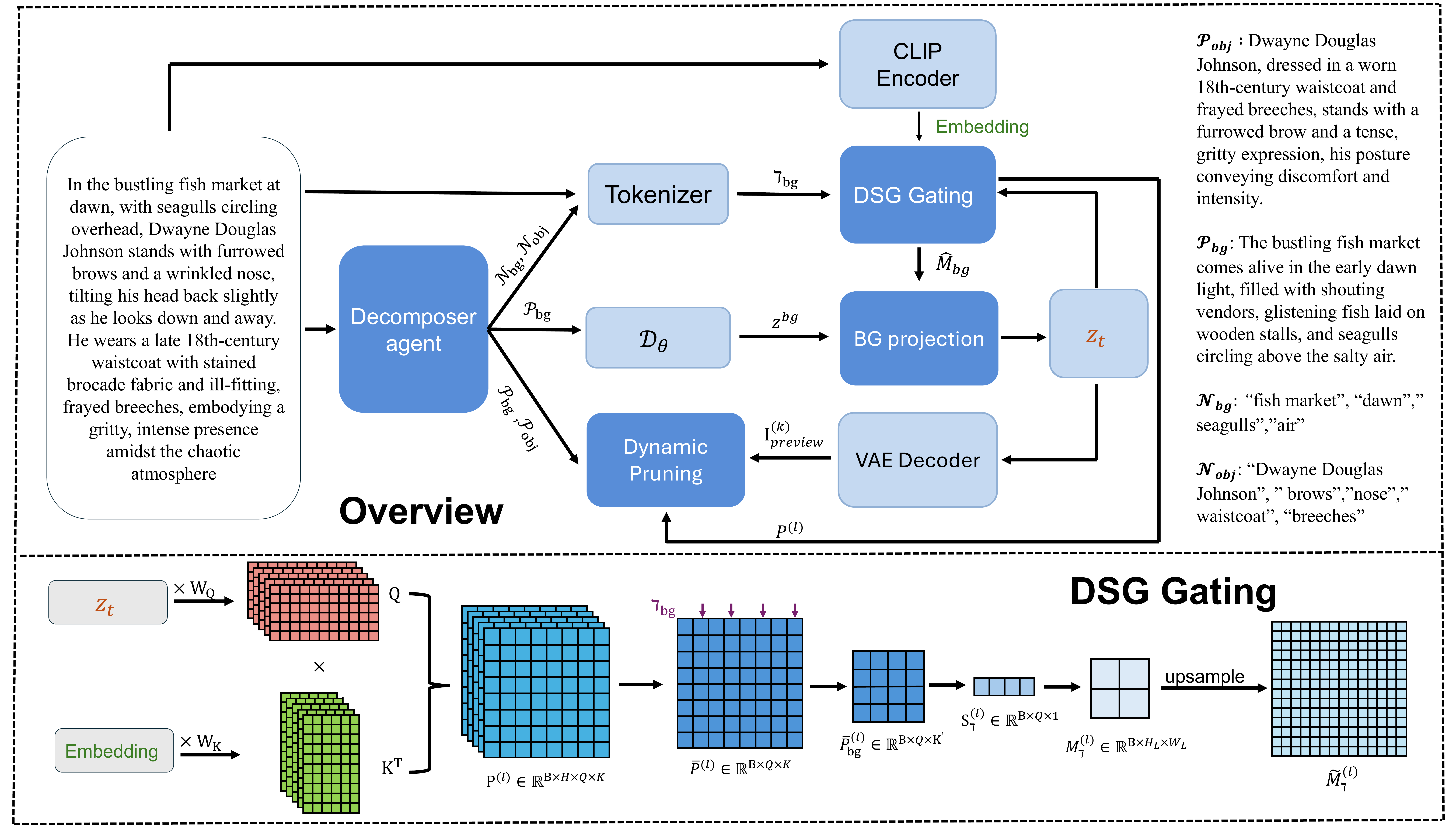}

    \caption{
        \textbf{Overview of the proposed framework.} 
        We expect a long text prompt as input. The \textbf{Decomposer Agent} parses it into structured components including background and object sentences ($\mathcal{P}_{\text{bg}}, \mathcal{P}_{\text{obj}}$) and their corresponding entity sets ($\mathcal{N}_{\text{bg}}, \mathcal{N}_{\text{obj}}$). 
        These sets are processed by the \textbf{Tokenizer} and embedded for diffusion conditioning \deng{\textbf{\( \mathcal{D}_\theta \)}}. 
        During denoising, the \textbf{Dynamic Spatial Guidance (DSG) Gating} module derives a time-varying background soft mask $\widehat{M}_{\text{bg}}$ from cross-attention statistics over background tokens, as illustrated in the lower panel. 
        This mask guides the \textbf{Background Projection (BG)} step [Eq.~\eqref{eq:projection}], which softly anchors early latent states $z_t$ toward a reference background latent $z_t^{\text{bg}}$, ensuring global consistency without retraining. 
        Multiple stochastic trajectories are explored in parallel, and the \textbf{Dynamic Pruning} module periodically evaluates each path using attention-tail entropy and CLIP alignment scores to discard incoherent candidates. 
        The bottom subfigure details the computation flow of DSG Gating, including attention aggregation, token grouping, and spatial upsampling for mask fusion.
        } 
        \vspace{-9mm}
    % \caption{Overview of the proposed framework. }
    \label{fig:pipeline}
\end{figure*}

We propose a training-free framework for achieving Object-Background Compositionality in text-to-image diffusion,
capable of parsing long prompts that describe both foreground and background.
The overall framework comprises four components: (i) a decomposer agent parses each freeform prompt into four elements \yl{(foreground and background sentences and entities)}, (ii) DSG Gating soft background masking, (iii) scheduled background projection in latent space, and (iv) multi-path exploration with quality-guided pruning. 
%Each component is defined formally in the following subsections.
Our method, illustrated in Fig.~\ref{fig:pipeline}, applies \textbf{Dynamic Spatial Guidance (DSG)} to extract background-aware attention maps $\mathbf{M}_{\text{bg}}$, enabling Latent-space \textbf{Background Projection} that anchors global scene structure. 
To maintain generative diversity while improving structural coherence, multiple diffusion trajectories are explored in parallel and adaptively pruned by a \textbf{Dynamic Pruning} module that combines attention-tail entropy $\mathcal{H}_{\text{Tail}}$ and CLIP alignment $\mathcal{S}_{\text{CLIP}}$. 
% \vspace{-6mm}
\subsection{Prompt Decomposition and Validation}
\vspace{-2mm}
\label{Long Prompt Decomposition}
We introduce a \emph{Decomposer Agent} that formally transforms the freeform prompt into four structured components: (i) a dedicated background sentence $\mathcal{P}_{\text{bg}}$, (ii) a main object sentence $\mathcal{P}_{\text{obj}}$, 
(iii) an entity set $\mathcal{N}_{\text{obj}}$ for main object control, and (iv) an entity set $\mathcal{N}_{\text{bg}}$ for background elements. Candidate decompositions are validated by a scoring function that enforces verbatim presence of all extracted nouns within the composition. This dual-stage \emph{composition–decomposition cycle} provides the semantic priors ($\mathcal{N}_{\text{obj}}, \mathcal{N}_{\text{bg}}$) required to ground our Cross-Attention masks for downstream control.
% \vspace{-5mm}
\subsection{Object-Background Decoupling with Dynamic Spatial Guidance Gating} %(DSG Gating)}
% \vspace{-4mm}
\paragraph{Preliminaries}Let \(x_0\in\mathbb{R}^{C\times H\times W}\) denote an image and \(z_0\in\mathbb{R}^{C_z\times h\times w}\) its corresponding latent under the VAE, where typically \(h=H/s\) and \(w=W/s\) with scale factor \(s\). A latent diffusion model (LDM) iteratively denoises latents using a denoising network \( \mathcal{D}_\theta \) conditioned on text embeddings. The diffusion scheduler provides a sequence of timesteps \(\{t_1,\dots,t_T\}\) and cumulative products \(\alpha_t\). For each timestep \(t\), the forward noising of a clean latent \(z_0\) is
\begin{equation}
z_t = \sqrt{\alpha_t}\, z_0 + \sqrt{1-\alpha_t}\,\epsilon,\quad \epsilon\sim\mathcal{N}(\mathbf{0},\mathbf{I}).
\label{eq:zt_noising}
\end{equation}

We denote by \(P^{(l)}\in\mathbb{R}^{B\times H^{(l)}\times Q^{(l)}\times K}\) the cross-attention probability tensor recorded at layer \(l\) within the denoising network $\mathcal{D}_\theta$, where \(B\) is batch size, \(H^{(l)}\) the number of attention heads, \(Q^{(l)}\) the number of spatial queries (tokens corresponding to spatial latent positions), 
and \(K\) the number of text tokens. 

We assume access to the tokenizer and the tokenized prompt so that we can specify two index sets of token indices \(\mathcal{T}_{\mathrm{obj}}\) and \(\mathcal{T}_{\mathrm{bg}}\) that represent object-related and background-related tokens respectively (formed by matching token text to pre-defined semantic word lists).

\begin{figure}[!b]
    \centering
    \includegraphics[width=\linewidth]{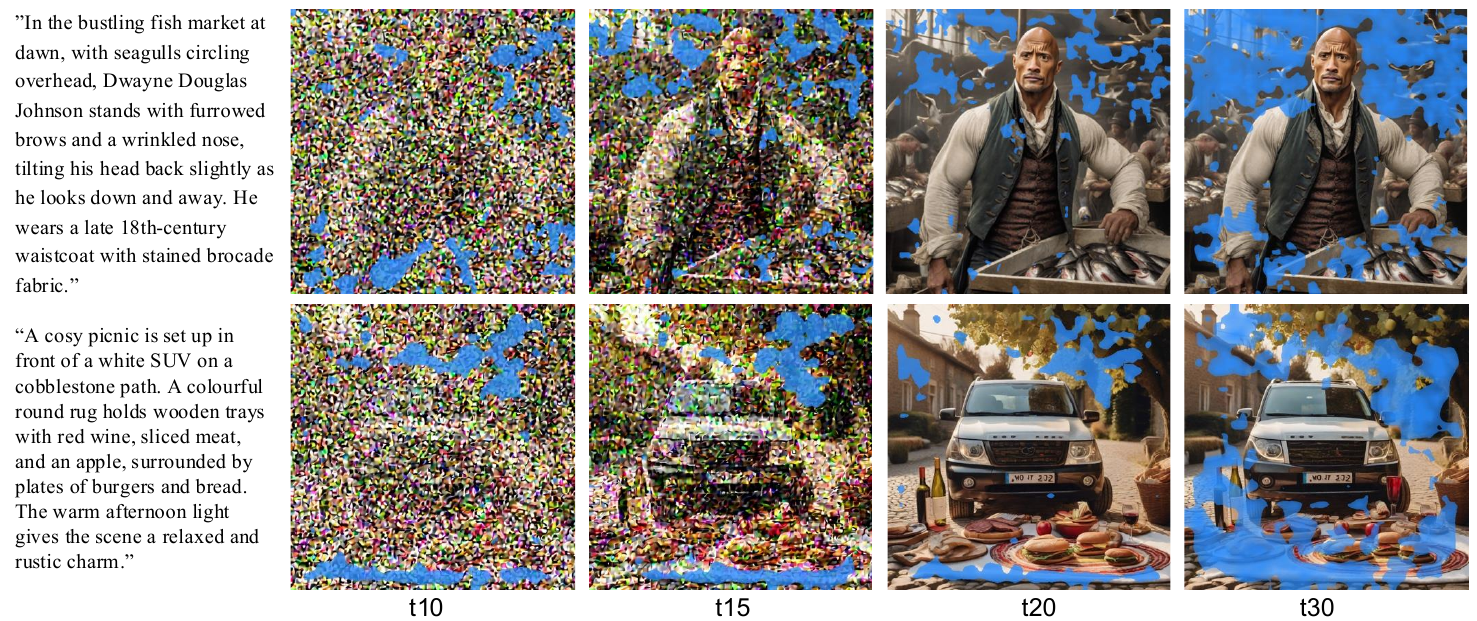}
    \vspace{-2mm}
    \caption{Dynamic Evolution of the Background Soft Mask ($\widehat{M}_{\mathrm{bg}}$) during Latent Denoising. The mask is computed dynamically at each step by aggregating cross-attention over background tokens ($\mathcal{T}_{\mathrm{bg}}$).}
    \label{fig:dsg_mask_evolution}
\end{figure}
\vspace{-3mm}
\paragraph{DSG Gating.} \yd{Several recent methods use attention or mask based guidance for region control, but their masks are typically static~\cite{wang2024compositional,chen2024region} or layer specific~\cite{chang2025maskattn}. In contrast, our DSG derives soft latent masks directly from internal multi-layer attentions, enabling dynamic and temporally coherent object-background decoupling. 
Specifically,} for each recorded layer \(l\) we average over heads to obtain a query-to-token matrix:
\begin{equation}
\bar{P}^{(l)}(b,q,k) = \frac{1}{H^{(l)}}\sum_{h=1}^{H^{(l)}} P^{(l)}(b,h,q,k),\quad b\in\{1,\dots,B\}.
\label{eq:avg_heads}
\end{equation}
Given a token-set \(\mathcal{T}\subseteq\{1,\dots,K\}\) (either \(\mathcal{T}_{\mathrm{obj}}\) or \(\mathcal{T}_{\mathrm{bg}}\)), we aggregate across those token indices to obtain a spatial attention map per query:
\begin{equation}
S^{(l)}_{\mathcal{T}}(b,q) \;=\; \frac{1}{|\mathcal{T}|}\sum_{k\in\mathcal{T}} \bar{P}^{(l)}(b,q,k).
\label{eq:agg_tokenset}
\end{equation}
We reinterpret the query index \(q\in\{1,\dots,Q^{(l)}\}\) as a spatial lattice of size \(h_l\times h_l\), where \(h_l = \sqrt{Q^{(l)}}\) is the side length of the feature map at layer \(l\). We then reshape \(S^{(l)}_{\mathcal{T}}\) to spatial maps \(M^{(l)}_{\mathcal{T}}(b)\in\mathbb{R}^{1\times h_l\times h_l}\). Each map is bilinearly upsampled to the target latent resolution \((h,w)\) producing \(\widetilde{M}^{(l)}_{\mathcal{T}}(b)\in\mathbb{R}^{1\times h\times w}\). Finally we fuse across all \(L\) layers with simple averaging followed by min-max normalization:
\begin{equation}
\widehat{M}_{\mathcal{T}}(b) \;=\; \mathrm{norm}_{[0,1]}\!\left(\frac{1}{L}\sum_{l=1}^{L}\widetilde{M}^{(l)}_{\mathcal{T}}(b)\right),
\label{eq:mask_fusion}
\end{equation}
where \(L\) is the total number of cross-attention layers used for fusion. The operation 
\(\mathrm{norm}_{[0,1]}(X) = \frac{X-\min (X)}{\max (X)-\min (X) + \varepsilon}\) has \(\varepsilon\) being a small constant for numerical stability. When only one of \(\mathcal{T}_{\mathrm{obj}}\) or \(\mathcal{T}_{\mathrm{bg}}\) is non-empty, the other is taken as the complement: \(\widehat{M}_{\mathrm{bg}} = 1-\widehat{M}_{\mathrm{obj}}\).

The masks \(\widehat{M}_{\mathrm{obj}}\) and \(\widehat{M}_{\mathrm{bg}}\) are \emph{soft} spatial masks in latent resolution that represent the model's internal soft assignment of each spatial query with object or background semantics. The masks are computed dynamically across the denoising trajectory and the internal spatial focus evolves over time, as depicted in Fig.~\ref{fig:dsg_mask_evolution}.
\vspace{-4mm}
\subsection{Latent-space Background Soft Projection} %(BG projection)}
\yd{Let \(z_t\) be the current latent at timestep \(t\) and \(z^{\mathrm{bg}}_t\) a pre-generated background latent. We apply a convex projection at early sampling stages to softly align background regions:}
\begin{equation}
\label{eq:projection}
z_t' \leftarrow z_t + \lambda(t)\,\widehat{M}_{\mathrm{bg}}\odot\bigl(z^{\mathrm{bg}}_t - z_t\bigr).
\end{equation}
% where \(\odot\) denotes channel-wise broadcasted multiplication, and \(\lambda(t)\in[0,1]\) is a scheduling coefficient that depends on the relative sampling stage. In our implementation \(\lambda(t)\) decays linearly to zero after an early-phase ratio \(r_{\max}\):
\yd{Here \(\odot\) denotes elementwise multiplication and \(\lambda(t)\in[0,1]\) is a decaying weight:}
\begin{equation}
\lambda(t) = 
\begin{cases}
\Lambda\left(1 - \dfrac{r(t)}{r_{\max}}\right) & r(t) < r_{\max},\\[4pt]
0 & r(t)\ge r_{\max},
\end{cases}
\qquad r(t)=\dfrac{\text{step}(t)}{T-1},
\label{eq:lambda_schedule}
\end{equation}
where \(\Lambda\) is a user-set maximum blend, and \(r_{\max}\) defines the boundary of the background dominance stage. \yd{Using Eq.~\eqref{eq:projection}, the projection blends \(z_t\) toward \(z^{\mathrm{bg}}_t\) on background regions during early steps, promoting coherent background formation without altering foreground dynamics.}
% \vspace{-4mm}
\begin{figure*}[t]
    \centering
    \includegraphics[width=\linewidth]{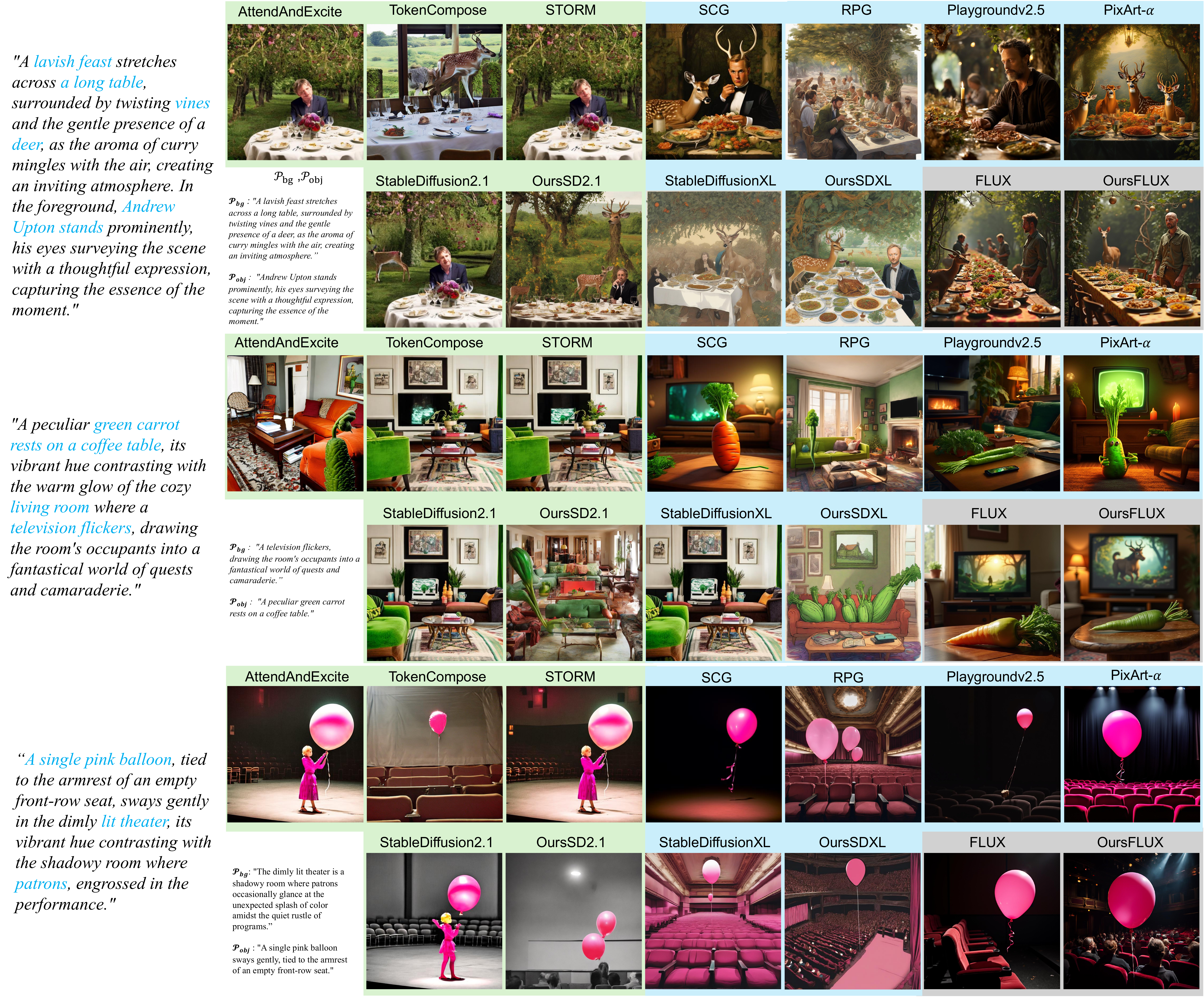}
    % \vspace{-5mm}
    \caption{
        Qualitative comparison across diffusion backbones. 
        Methods are grouped by backbone type, highlighted with colored blocks: 
        \textcolor{green!50!black}{\textbf{green}} for SD2.1-based models, 
        \textcolor{blue!60!black}{\textbf{blue}} for SDXL/DiT-based models, and 
        \textcolor{gray!70!black}{\textbf{gray}} for FLUX-based models. 
        Each row depicts images generated from the same prompt and random seed. 
        }
    \label{fig:twoRowsResults}
    % \vspace{-8mm}
\end{figure*}

\subsection{Dynamic Pruning Mechanism}
\label{Dynamic Pruning Mechanism}
\yd{Inspired by confidence estimation in LLM reasoning~\cite{razghandi2025cer,zhou2025bridging,fu2025deep}, 
which estimates the reliability of reasoning paths within the textual output space using predictive uncertainty or log-probability signals, 
we extend this idea to the \emph{cross-modal generative domain.}}

To explore the stochastic latent manifold, we instantiate \(K\) trajectories with distinct seeds and initial latents \(\{z_0^{(k)}\}_{k=1}^K\), each maintaining its own scheduler and prompt embeddings. To save memory, the denoising network \(\mathcal{D}_\theta\)  is applied sequentially across trajectories per timestep, matching parallel evaluation while requiring only a single network in memory. The background projection in Eq.~\eqref{eq:projection} is then performed independently for each \(z_t^{\mathrm{bg},(k)}\).

We propose a dynamic path-quality score combining an internal attention-based statistic (tail entropy) with an external semantic metric (CLIP similarity). The overall process, including $\text{DSG}$ correction and quality-driven pruning, is summarized in Algorithm~\ref{alg:main}.
\begin{figure*}[t]
    \centering
    \includegraphics[width=\linewidth]{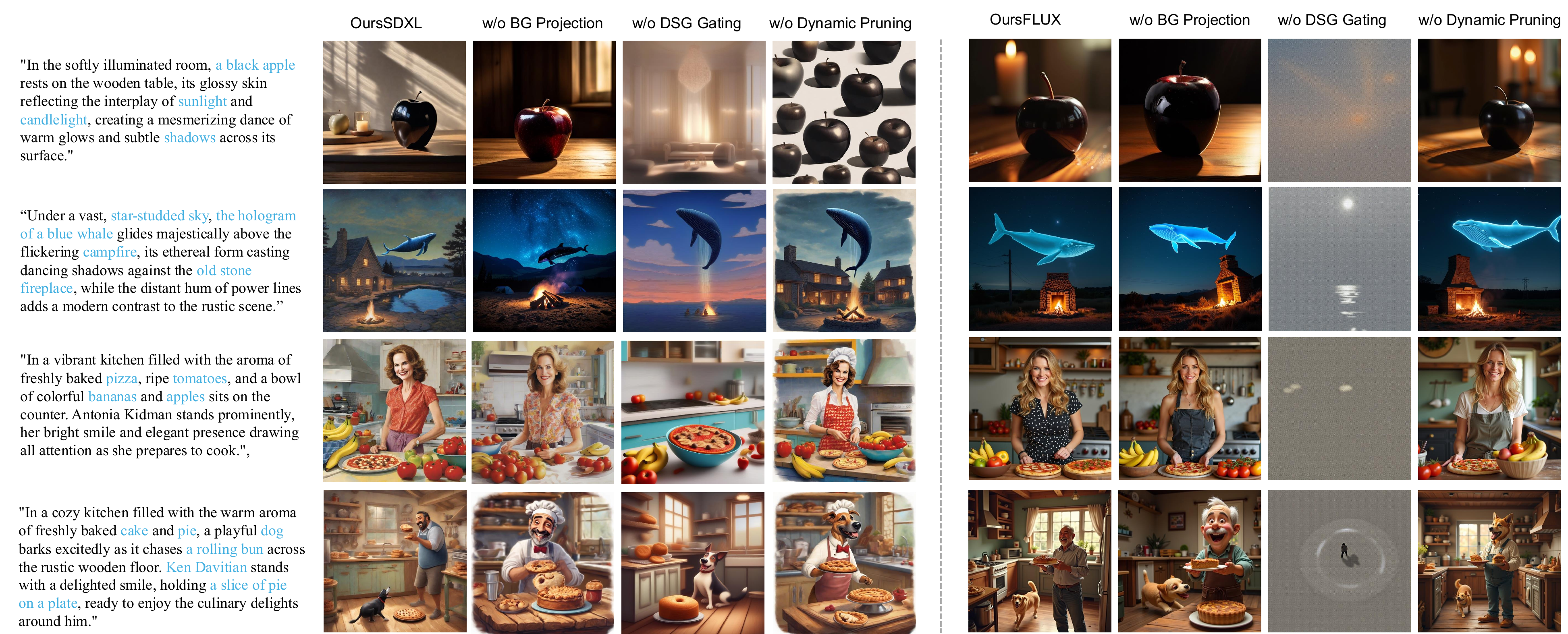}
    \caption{Ablation study on key components. We validate the necessity of the $\text{DSG}$ framework on both $\text{SDXL}$~\cite{podell2023sdxl} and $\text{FLUX}$~\cite{flux2024} architectures. \textbf{(A) w/o Projection:} Removing $\text{BG}$ $\text{Projection}$ causes the model to ignore background keywords ($\mathcal{P}_{\mathrm{bg}}$). \textbf{(B) w/o DSG Gating:} Omitting the $\text{DSG}$ soft mask (i.e., using $\widehat{M}_{\mathrm{bg}} \equiv 1$) forces a global correction, leading to catastrophic blurring and structural ruin, especially in $\text{FLUX}$~\cite{flux2024}. Spatial gating is mandatory for safe latent modification. \textbf{(C) w/o Dynamic Pruning:} Disabling $\text{Online Pruning}$ allows paths with high $\text{Attention}$ $\text{Entropy}$ to proceed, frequently resulting in \textbf{Concept Collapse} and poor semantic focus.}
    % \vspace{-5mm}
\label{fig:ablation_study}
\end{figure*}
\vspace{-2mm}
\paragraph{Tail entropy} For each path, the attention matrix $P \in \mathbb{R}^{Q \times K}$ used for entropy computation is obtained by averaging raw attention tensors $P^{(l)}$ across all $\text{Cross-Attention}$ layers, heads, and batches within the pruning window, capturing the overall query-token attention pattern. The pixel-to-token entropy for each spatial query $q$ is then defined as:
\begin{equation}
H_{\mathrm{pix}\to\mathrm{tok}}(q) \;=\; -\sum_{k=1}^K p_{qk}\log p_{qk}.
\label{eq:H_pix2tok}
\end{equation}
Token-to-pixel entropy after row-normalization of \(P^\top\) is defined similarly:
\begin{equation}
H_{\mathrm{tok}\to\mathrm{pix}}(k) \;=\; -\sum_{q=1}^Q a_{kq}\log a_{kq},
\quad a_{kq}=\frac{p_{qk}}{\sum_{q'} p_{q'k}}\;.
\label{eq:H_tok2pix}
\end{equation}
\yang{Cross-attention encodes asymmetric correspondences between image queries and text tokens~\cite{hertz2022prompt}. We therefore compute entropy in both directions: \(H_{\mathrm{pix}\to\mathrm{tok}}\) quantifies how selectively pixels attend to concepts, while \(H_{\mathrm{tok}\to\mathrm{pix}}\) captures how spatially coherent each concept is realized~\cite{tumanyan2023plug}.}

\deng{Focusing} on the low-entropy tail highlights the most confident, semantically grounded correspondences. We define a \emph{tail-entropy score} to quantify attention concentration. 
For each path, we compute the mean entropy over the top-$r$ fraction of values in both pixel-to-token and token-to-pixel directions:
\begin{equation}
\mathrm{TailH}=\tfrac{1}{2}\!\left[
\frac{1}{\lceil r n\rceil}\!\sum_{i=1}^{\lceil r n\rceil}\! H^{\mathrm{pix}\to\mathrm{tok}}_{(i)}+
\frac{1}{\lceil r n\rceil}\!\sum_{i=1}^{\lceil r n\rceil}\! H^{\mathrm{tok}\to\mathrm{pix}}_{(i)}
\right],
\label{eq:TailH}
\end{equation}
where $H_{(i)}$ denotes entropies sorted in descending order and $n$ is the number of elements. 
A lower $\mathrm{TailH}$ indicates more concentrated attention and empirically correlates with clearer semantic focus.

\paragraph{Semantic Alignment}
\deng{Complementing the internal uncertainty signal, we use a semantic alignment signal from preview images, instantiated with a CLIP-family model but replaceable by other vision-language models.} For \deng{each} path \(k\), the current latent \(z_t^{(k)}\) is decoded to a low-resolution preview image \(\mathbf{I}_{\mathrm{preview}}^{(k)}\). 

\deng{In our implementation, we use CLIP~\cite{radford2021learning} to compute a clause-level semantic score. Specifically, we measure the mean cosine similarity between the image embedding $F_{\mathrm{img}}$ and the embeddings of prompt clauses $\mathcal{P}_{\mathrm{clauses}}$, obtained by decomposing the full prompt into object and background related descriptions.} 
% defined as the mean cosine similarity between the image embedding $F_{\mathrm{img}}$ and the embeddings of the set of prompt clauses $\mathcal{P}_{\mathrm{clauses}}$:
\begin{equation}
\mathrm{CLIP}^{(k)} = \frac{1}{|\mathcal{P}_{\mathrm{clauses}}|} \sum_{c \in \mathcal{P}_{\mathrm{clauses}}} \cos(F_{\mathrm{img}}(\mathbf{I}_{\mathrm{preview}}^{(k)}), F_{\mathrm{txt}}(c)).
\label{eq:clip_score}
\end{equation}
We choose an embedding-based alignment signal as it enables relative comparison between multiple noisy preview images at early timesteps, which is critical for pruning decisions before full image formation.
\deng{Moreover, CLIP-family models are commonly used as semantic backbones in diffusion pipelines, making them a practical and stable choice in this setting.
To assess sensitivity to the specific model choice, we also replace CLIP with OpenCLIP \yl{trained with different data}, and observe highly consistent  (e.g., CLIP\_I: 0.3538 vs. 0.3520; NounRecall (also see \yl{Sec.}~\ref{evaluation}): 0.5884 vs. 0.5911), indicating \yl{reliable prune behavior} and final performance.}
For each path \(k\) we maintain recent windows of \(\{\mathrm{TailH}^{(k)}\}\) and \(\{\mathrm{CLIP}^{(k)}\}\). The two signals are min-max normalized independently over the current live-path set, and a composite score is formed:
\begin{equation}
S^{(k)} = w_{\mathrm{tail}}\cdot\mathrm{norm}\bigl(-\mathrm{TailH}^{(k)}\bigr) + w_{\mathrm{clip}}\cdot\mathrm{norm}\bigl(\mathrm{CLIP}^{(k)}\bigr),
\label{eq:score}
\end{equation}
where \(w_{\mathrm{tail}},w_{\mathrm{clip}}\) are positive weights and \(\mathrm{norm}\) denotes min-max scaling to \([0,1]\). Paths with \(S^{(k)} < \tau_q\) are pruned at each interval \deng{$I$}, provided they satisfy the stability criterion of having a minimum recorded window length\yd{, where $\tau_q$ denotes the $q$-quantile of $\{S^{(k)} \mid k \in \mathcal{A}_t\}$, 
serving as an adaptive pruning threshold for the current live-path set.}
\begin{algorithm}[t]
\caption{Multi-path sampling with dynamic pruning}
\label{alg:main}
\begin{algorithmic}[1]
\Require Prompt $p$, seeds $\{s_k\}_{k=1}^K$, steps $T$, pruning interval $I$
\State Initialize $K$ paths with latents $z_0^{(k)}$ and background latents $z_0^{\mathrm{bg},(k)}$
\For{each timestep $t \in \{1,\dots,T\}$}
  \For{each live path $k$ sequentially}
    \State Compute noise prediction $\hat{\epsilon}^{(k)}$ via UNet with text conditioning
    \State Form guided prediction with CFG and perform scheduler step to obtain $z_t^{(k)}$
    \State Collect cross-attention $P^{(l)}$ into controller and compute $\widehat{M}_{\mathrm{bg}}^{(k)}$
    \If{$t$ is in the background stage}
      \State Apply projection $z_t^{(k)} \leftarrow z_t^{(k)}+\lambda(t)\,\widehat{M}_{\mathrm{bg}}^{(k)}\odot\bigl(z_t^{\mathrm{bg},(k)}-z_t^{(k)}\bigr)$
    \EndIf
  \EndFor
  \If{$t$ is a pruning interval \deng{$I$}}
    \State Decode previews, compute $\mathrm{TailH}$ and $\mathrm{CLIP}$ per path, update composite scores $S^{(k)}$
    \State Prune paths with $S^{(k)}$ below quantile threshold $\tau_q$
  \EndIf
\EndFor
\State Decode remaining latents to produce final images.
\end{algorithmic}
\end{algorithm}
\begin{table*}[!t]
    \centering
    \small
    \renewcommand{\arraystretch}{1.05}
    \caption{Global mean metrics comparison across methods. \textbf{Bold} values indicate the best performance within each group, and underlined values denote the second best.}
    \resizebox{\textwidth}{!}{
    \begin{tabular}{lcccccc}
    \toprule
    Method & Venue & CLIP\_I~\cite{radford2021learning} & BLIP~\cite{li2022blip} & NounRecall & HPSv3~\cite{ma2025hpsv3widespectrumhumanpreference} & ID-Sim~\cite{deng2019arcface} \\
    \midrule
    \rowcolor{gray!10}
    \multicolumn{7}{l}{\textbf{SD2.1-based}} \\
    AttendAndExcite~\cite{chefer2023attend} & SIGGRAPH'23 & \underline{0.3295} & 0.4535 & \underline{0.5567} & \underline{8.1081} & - \\
    TokenCompose~\cite{wang2024tokencompose} & CVPR'24 & 0.3258 & \underline{0.4676} & 0.5509 & 6.4276 & - \\
    STORM~\cite{han2025spatial} & CVPR'25 & 0.3278 & 0.4586 & 0.5527 & \textbf{8.2355} & - \\
    SD2.1~\cite{rombach2022high} & CVPR'22 & 0.3278 & 0.4405 & 0.5326 & 5.6710 & - \\
    OursSD2.1 & Ours & \textbf{0.3364} & \textbf{0.4716} & \textbf{0.5586} & 6.9645 & - \\
    \rowcolor{gray!10}
    \multicolumn{7}{l}{\textbf{SDXL/DiT-based}} \\
    
    PixArt-$\alpha$~\cite{chen2023pixart} & ICLR'24 & 0.3499 & 0.5008 & 0.5755 & 8.4084 & 0.041 \\
    playgroundv2.5~\cite{li2024playground} & arXiv'24 & \underline{0.3510} & \underline{0.4822} & \underline{0.5850} & \underline{9.2906} & \underline{0.112} \\
    SCG~\cite{wang2025towards} & CVPR'25 & 0.3479 & 0.5016 & 0.5769 & \textbf{10.6348} & 0.006 \\
    RPG~\cite{yang2024mastering} & ICML'24 & 0.3380 & 0.4961 & 0.5683 & 7.4521 & 0.105 \\
    SDXLBaseline~\cite{podell2023sdxl} & ICLR'24 & 0.3412 & 0.4877 & 0.5534 & 7.2251 & 0.063 \\
    OursSDXL & Ours & \textbf{0.3538} & \textbf{0.5091} & \textbf{0.5884} & 8.8063 & \textbf{0.130} \\
    
    \rowcolor{gray!10}
    \multicolumn{7}{l}{\textbf{Flux-based}} \\
    
    FLUX~\cite{flux2024} & GitHub'24 & \underline{0.3420} & \underline{0.4952} & \underline{0.6115} & \underline{11.1425} & \underline{0.010} \\
    OursFLUX & Ours & \textbf{0.3520} & \textbf{0.5152} & \textbf{0.6389} & \textbf{11.9849} & \textbf{0.040} \\
    
    \bottomrule
    \end{tabular}
    }
    \label{tab:global_metrics}
    \vspace{-6mm}
\end{table*}
% \vspace{-6mm}
\section{Experiments}
% \vspace{-4mm}
\paragraph{Implementation Details}
Following standard practice, we implement our framework on \textbf{Stable Diffusion~2.1}~\cite{rombach2022high} and \textbf{Stable Diffusion~XL}~\cite{podell2023sdxl} for the main experiments, and additionally extend it to \textbf{FLUX}~\cite{flux2024}. 
However, since the baseline quality of SD~2.1 is substantially lower, it becomes difficult to disentangle the effect of our modules from the limitations of the underlying model. 
Therefore, our ablation studies focus on SDXL~\cite{podell2023sdxl} and FLUX~\cite{flux2024}, where the base generation quality is more reliable. 
All experiments are conducted on an NVIDIA RTX~4090 GPU for SD2.1/XL and \yl{DiT}-based methods, and on a NVIDIA H100 GPU for FLUX-based models. 
\subsection{Evaluation Dataset}
\vspace{-3mm}
\yl{As there are no existing benchmarks for object-background compositional T2I, }we create a structured evaluation dataset %for the Object-Background Compositional task, 
as follows.
We derive a vocabulary of entities from the T2I-CompBench~\cite{huang2023t2i}. 
All captions are tokenized and part-of-speech tagged using spaCy, retaining only nouns and proper nouns as a base vocabulary $\mathcal{V}$. 
We embed all words via a sentence-transformer model~\cite{reimers2019sentence} and build a $k$-nearest \yl{neighbor} graph under cosine similarity. 
Applying Louvain community detection~\cite{blondel2008fast} yields coherent \textit{scene clusters}, each representing a semantically consistent context (e.g., \textit{``market–vendor–fish–basket''}). 
Cluster centroids are used to extract representative background entities, forming a compact, interpretable set of scene prototypes.
To generate complex text prompts for evaluation, each scene \yang{prototype} is paired with a foreground entity sampled from two domains: $100$ celebrities from VGGFace2~\cite{cao2018vggface2} and $200$ non-person entities from T2I-CompBench~\cite{huang2023t2i}. 
Each pair $\{(\text{Nouns}_{\text{BG}}, \text{Nouns}_{\text{OBJ}})\}$ is fed into a constrained LLM-based Synthesis Agent (GPT-4o~\cite{achiam2023gpt}) that produces a single integrated prompt \deng{that both describes} the scene ambience and \deng{specifies} the main subject. 
Full details of vocabulary construction, graph clustering, and sampling parameters are provided in the Supplementary Material.
 
\vspace{-5mm}
\subsection{Evaluation Results}
\label{evaluation}
\vspace{-3mm}
We conduct both qualitative and quantitative evaluations to assess the effectiveness of our proposed framework. 
Comparisons are made against standard diffusion baselines~\cite{rombach2022high,podell2023sdxl,flux2024} and recent state-of-the-art compositional generation methods~\cite{chefer2023attend,yang2024mastering,han2025spatial,wang2024tokencompose,chen2023pixart,li2024playground,wang2025towards}. 
\vspace{-3mm}
\paragraph{Qualitative Comparison.}
Figure~\ref{fig:twoRowsResults} shows representative images grouped by backbone type. Each row shows images generated from the same prompt \yd{and seed.} 
Across all groups, our method consistently produces scenes where the main object and its surrounding environment are jointly rendered with superior semantic and visual coherence, most closely aligning with the text description.
Among all methods, our method yields more semantically consistent and visually coherent scenes. Notably, the FLUX-based variant~\cite{flux2024} produces the most faithful foregrounds and complete backgrounds.
Methods such as Attend-and-Excite~\cite{chefer2023attend}, TokenCompose~\cite{wang2024tokencompose}, and STORM~\cite{han2025spatial}, which rely on the Stable Diffusion2.1~\cite{rombach2022high} backbone, struggle to produce artifact-free images and often fail to preserve key objects or background structures. 
Our \textit{OursSD2.1}, also built upon Stable Diffusion~2.1~\cite{rombach2022high}, attempts to enhance the completeness of background elements but is subject to the limitations of the base model, occasionally introducing visual artifacts. 
In contrast, diffusion transformer (DiT)-based methods such as PixArt\text{-}alpha~\cite{chen2023pixart} and playgroundv2.5~\cite{li2024playground} produce visually pleasing results; however, they tend to ignore the main object or omit background elements mentioned in the prompt. 
\textit{OursSDXL} achieves the best balance, successfully maintaining the semantic integrity of both the main subject and contextual details. 

\paragraph{Quantitative Evaluation}
Table~\ref{tab:global_metrics} reports quantitative results across all groups. 
We evaluate semantic alignment, compositional grounding, perceptual quality, and identity consistency using CLIP-I~\cite{radford2021learning}, BLIP~\cite{li2022blip}, HPSv3~\cite{ma2025hpsv3widespectrumhumanpreference}, and ID-Sim based on ArcFace~\cite{deng2019arcface}. 
To explicitly assess structural grounding, we introduce \textbf{NounRecall}, which quantifies whether foreground and background entities specified in the prompt are correctly detected by an open-vocabulary detector~\cite{minderer2022simple}. 
For identity evaluation, images without a valid detected face are assigned a penalty score of $-1$, enforcing strict handling of practical failure cases. 
Full metric definitions are provided in the supplementary material~\ref{metrics}.

Our method consistently achieves \deng{competitive} performance \deng{across metrics related to} structural fidelity and semantic disentanglement. 
OursSDXL and OursSD2.1 yield the highest CLIP\_I scores~\cite{radford2021learning} (0.3538 and 0.3364), demonstrating the effectiveness of our framework in maintaining joint object and background consistency. 
OursFLUX attains the best NounRecall (0.6389) and BLIP~\cite{li2022blip} (0.5152), confirming that structured priors promote faithful entity grounding. 
Furthermore, OursSDXL achieves the highest ID\_Sim~\cite{deng2019arcface} (0.130) among SDXL and \yl{DiT}-based models, showing that quality-driven pruning enhances subject preservation. 
Notably, OursSD2.1 and OursSDXL do not lead in the aesthetic metric. As HPSv3~\cite{ma2025hpsv3widespectrumhumanpreference} ranks preferences by human aesthetic preference, it naturally reflects human bias toward salient foreground subjects rather than holistic scene composition~\cite{borji2015reconciling,misra2016seeing}.
Even under this bias, our methods \yl{with no explicit optimization for aesthetics} still outperform the base models (SD2.1 and SDXL), showing better background–foreground balance. 

In addition, FLUX models exhibit relatively lower CLIP\_I scores~\cite{radford2021learning} despite strong qualitative alignment (Fig.~\ref{fig:twoRowsResults}) These observations motivated us to conduct a user study to further assess human-perceived semantic fidelity and visual balance.
\begin{table}[H]
    \centering
    \scalebox{1.0}{
    \begin{tabular}{lccc}
    \toprule
    \textbf{Method} & \textbf{Full} & \textbf{Background} & \textbf{Object} \\
    \midrule
    \rowcolor{gray!10}
    \multicolumn{4}{l}{\textbf{SD2.1-based models}} \\
    AttendAndExcite~\cite{chefer2023attend}      & \textbf{0.3472} & 0.3782 & \textbf{0.4456} \\
    TokenCompose~\cite{wang2024tokencompose}     & 0.2412 & 0.2647 & 0.3706 \\
    STORM~\cite{han2025spatial}                  & 0.3091 & 0.3697 & 0.3212 \\
    Stable Diffusion 2.1~\cite{rombach2022high}  & 0.2623 & \underline{0.3934} & 0.3770 \\
    OursSD2.1                                   & \underline{0.3121} & \textbf{0.4162} & \underline{0.3815} \\
    \rowcolor{gray!10}
    \multicolumn{4}{l}{\textbf{SDXL and DiT-based models}} \\
    PixArt\text{-}alpha~\cite{chen2023pixart}    & 0.6182 & \underline{0.6303} & \underline{0.6000} \\
    playgroundv2.5~\cite{li2024playground}       & 0.6136 & 0.5625 & \textbf{0.6136} \\
    SCG~\cite{wang2025towards}                       & \underline{0.6301} & 0.5959 & 0.5822 \\
    RPG~\cite{yang2024mastering}                 & 0.3876 & 0.3721 & 0.3721 \\
    Stable Diffusion XL~\cite{podell2023sdxl}    & 0.5054 & 0.4348 & 0.4783 \\
    OursSDXL                                     & \textbf{0.7159} & \textbf{0.6761} & 0.5852 \\
    \rowcolor{gray!10}
    \multicolumn{4}{l}{\textbf{Flux-based models}} \\
    FLUX~\cite{flux2024}                         & \underline{0.7640} & \underline{0.7205} & \underline{0.6708} \\
    OursFLUX                                    & \textbf{0.8284} & \textbf{0.7456} & \textbf{0.7633} \\
    \bottomrule
    \end{tabular}
    }
    \caption{User study results across all methods. Each participant compared pairs of generated images under three criteria: full description, background description, and main object description. Reported values denote win rates per method. Note that \emph{PixArt\text{-}alpha} and SCG \yl{adopt} a DiT architecture; and \emph{playgroundv2.5} follows the SDXL architecture but is \emph{trained from scratch}. \textbf{Bold} values indicate the best performance within each group, and \underline{underlined} values denote the second best.}
    \vspace{-12mm}
    \label{tab:userStudy}
\end{table}
% \vspace{-6mm}
\vspace{-2mm}
\paragraph{User Study}
Current metrics mainly focus on foreground fidelity and attribute accuracy, showing bias toward object-centric evaluation. To better assess foreground–background coherence, we conducted a controlled user study on human-perceived compositional quality. 
\deng{We randomly selected 12 prompts from our constructed dataset, sampling six from the object subset and six from the celebrity subset to ensure diversity across content types. For each sample, participants were shown the complete prompt along with its decomposed main-object and background descriptions, followed by images generated by two different methods. The two methods were randomly paired from the full set of evaluated approaches, such that any pair of methods could be compared; our method was not enforced to appear in every comparison.
} Each participant answered three forced-choice questions: (1) which image best matches the full prompt? (2) which best matches the background description? and (3) which best matches the main-object description? 

A total of 1,095 responses were collected from 30 users. As shown in Figure~\ref{tab:userStudy}, OursFLUX achieved the highest win rates across all criteria (0.83 / 0.75 / 0.76), surpassing all methods by a clear margin. SD2.1-based methods perform significantly worse in all categories, highlighting the dominant influence of the base model in training-free compositional control. \yang{While playgroundv2.5~\cite{li2024playground} and PixArt\text{-}alpha~\cite{chen2023pixart} achieve marginally higher scores on the main-object metric, this improvement arises from their model biases: PixArt\text{-}alpha’s DiT architecture concentrates attention on semantically salient regions, and playgroundv2.5’s training emphasizes aesthetic objects. Both favor subject-centric synthesis. In contrast, our SDXL-based framework retains the latent-space prior that governs scene coherence, leading to stronger background consistency and better overall prompt alignment.}
% \vspace{-4mm}
\deng{\paragraph{Inference Latency.} We report inference latency of our framework. On SD2.1, a standard generation takes $0.62$s per image, Attend-and-Excite~\cite{chefer2023attend} requires $14.51$s, STORM~\cite{han2025spatial} requires $9.316$s, TokenCompose~\cite{wang2024tokencompose} requires $7.57$s and our method takes $7.80$s. On SDXL, the baseline requires $3.82$s per image, the single-path RPG~\cite{yang2024mastering} takes $24.44$s, whereas our method takes $36.53$s. For Flux-based models, the standard Flux baseline takes $12.41$s per image, while our method requires $37.38$s.
Although our approach explores multiple trajectories, these paths are executed \emph{in parallel} during sampling, concentrating computation on promising candidates. Consequently, we obtain improved performance while maintaining a comparable inference time to prior compositional guidance methods.}
% \vspace{-5mm}
\subsection{Ablation Studies}
% \vspace{-3mm}
Ablation studies on both our \text{SDXL}-based and \text{FLUX}-based frameworks quantify the contribution of each integrated component. Some results are shown in Figure~\ref{fig:ablation_study}, with more in the supplementary materials.
% \vspace{-5mm}
\paragraph{$\text{w/o}$ Background Projection}
Removing the Latent-Space Background Projection compromises the model's ability to anchor the scene context. Without this early guided correction, the $\mathcal{D}_\theta$ network tends to ignore key background entities specified in $\mathcal{P}_{\mathrm{bg}}$, leading to generic or semantically inaccurate background regions.

% \vspace{-5mm}
\paragraph{$\text{w/o}$ DSG Gating}
The $\text{DSG}$ mechanism, which provides the $\widehat{M}_{\mathrm{bg}}$ soft mask, is critical for localization. When we omit the soft mask (i.e., treating $\widehat{M}_{\mathrm{bg}} \equiv 1$ or removing $\text{DSG}$ entirely), the $\text{BG}$ projection is forced to operate globally. This results in overly aggressive mixing across the entire latent space, including the foreground subject. For advanced architectures like \text{FLUX}~\cite{flux2024}, which uses multi-resolution streams, applying an unmasked global correction frequently ruins the entire image structure, leading to severe blurring or catastrophic concept collapse.
\vspace{-3mm}
\paragraph{$\text{w/o}$ Dynamic Pruning}
Finally, disabling the Dynamic Pruning mechanism leads to severe output variance. Without the mechanism to select for paths with stable internal attention ($\text{TailH}$) and high external semantic alignment ($\text{CLIP}$)~\cite{radford2021learning}, the parallel sampling trajectories are highly susceptible to Concept Collapse due to attention leakage or stochastic drift in early steps. The pruning component effectively acts as a quality filter, ensuring that the final output is drawn from the highest-fidelity compositional path.
% \vspace{-5mm}
\section{Conclusion}
% \vspace{-3mm}
\yang{
We introduce a training-free framework for \emph{Object-Background Compositionality}, balancing background and foreground rather than overly prioritizing foreground as is common in existing work. We employ Background Projection to decouple foreground and background generation, and manage integration with Dynamic Spatial Guidance (DSG) Gating, using a soft mask to harmonize the projected background structure with the developing subject. Recognizing high seed-dependent output variance, we propose Dynamic Pruning, sampling multiple paths concurrently; an internal attention-tail entropy ($\mathrm{TailH}$) provides semantic focus, and external CLIP alignment estimates prompt fidelity, allowing the system to select the most coherent final result. To facilitate evaluation in this under-explored area, we also contribute a structured Object–Background Compositionality evaluation dataset. \deng{Our method consistently improves performance across most metrics on SD2.1, SDXL, and FLUX, indicating stronger alignment with the overall prompt semantics.} %However, performance is heavily reliant on the base model, and our pruning cannot eliminate artifacts when all sampled paths are degraded. Also, our model prioritizes the presence of numerous complex background elements over perfect attribute binding for every single element.
}

% \clearpage\mbox{}Page \thepage\ of the manuscript. This is the last page.
% Now we have reached the maximum length of an ECCV \ECCVyear{} submission (excluding references and acknowledgements).
% References should start immediately after the main text, but can continue past p.\ 14 if needed. 
\clearpage  % TODO FINAL: This \clearpage needs to be removed from both review and camera-ready versions.

% \section*{Acknowledgements}
% Please insert your acknowledgments here.

\bibliographystyle{splncs04}
\bibliography{main}

\clearpage
\setcounter{page}{1}
\appendix
{\noindent {\huge Training-Free Object-Background Compositional T2I via Dynamic Spatial Guidance and Multi-Path Pruning (Appendix)}}
\section{Evaluation}
\subsection{Dataset}
\label{Evaluation Dataset}
To obtain a structured set of scene-level concepts, we first extract a vocabulary of entities from the T2I-CompBench dataset~\cite{huang2023t2i}. 
Given the raw text corpus $\mathcal{D}$, we employ the spaCy toolkit to tokenize and perform part-of-speech tagging \dy{(POS)}, and retain only common nouns (NOUN) and proper nouns (PROPN). 
Formally, for each token $t \in \mathcal{D}$ with POS tag $\mathrm{pos}(t)$, we define the noun set as
\dy{\begin{equation}
\mathcal{V} = \{\, t \mid \mathrm{pos}(t) \in \{\text{NOUN}, \text{PROPN}\} \,\}.%
\label{eq:vocab}%
\end{equation}}%
for foreground and background respectively. 
After deduplication and sorting, we obtain a vocabulary $\mathcal{V} = \{w_1, w_2, \dots, w_N\}$.

\paragraph{Semantic Embedding.}
Each noun $w_i$ is mapped into a dense representation $\mathbf{v}_i \in \mathbb{R}^d$ using a pre-trained sentence-transformer model (all-MiniLM-L6-v2), \dy{where $d$ denotes the embedding dimension.}
All vectors are $L_2$-normalized such that
\begin{equation}
    \mathbf{v}_i \leftarrow \frac{\mathbf{v}_i}{\|\mathbf{v}_i\|_2}, 
    \quad \forall i=1,\dots,N,
    \label{eq:normalize}
\end{equation}
so that cosine similarity reduces to the dot product.

\paragraph{k-NN Graph Construction.}
We build a $k$-nearest neighbor graph $\mathcal{G}=(\mathcal{V},\mathcal{E})$ based on cosine similarity. 
For each node $w_i$, we connect it to its top-$k$ most similar neighbors, provided their similarity exceeds a threshold \yd{$\eta$}. In practice, we set \dy{$\eta=0.5$}.
Specifically, the edge weight between $w_i$ and $w_j$ is defined as
\begin{equation}
    s_{ij} = \cos(\mathbf{v}_i, \mathbf{v}_j) = \mathbf{v}_i^\top \mathbf{v}_j,
    \quad (i,j) \in \mathcal{E}, \ s_{ij} \ge \yd{\eta}.
    \label{eq:similarity}
\end{equation}

\paragraph{Community Detection.}
To identify coherent semantic scenes, we apply the Louvain community detection algorithm on $\mathcal{G}$. 
The algorithm seeks a partition $\mathcal{C} = \{C_1,\dots,C_M\}$ that maximizes modularity $Q$, defined as
\begin{equation}
    Q = \frac{1}{2m} \sum_{i,j} \Big[ s_{ij} - \frac{k_i k_j}{2m} \Big] \, \delta(c_i, c_j),
    \label{eq:modularity}
\end{equation}
where $k_i=\sum_j s_{ij}$ is the degree of node $i$, $m=\tfrac{1}{2}\sum_{i,j} s_{ij}$ is the total edge weight, 
and $\delta(c_i, c_j)=1$ if $w_i$ and $w_j$ are assigned to the same community. 
Each community $C_m$ corresponds to a scene cluster.

\paragraph{Scene Representation.}
For each cluster $C_m$, we compute its centroid vector
\begin{equation}
    \boldsymbol{\mu}_m = \frac{1}{|C_m|} \sum_{i \in C_m} \mathbf{v}_i,
    \quad \boldsymbol{\mu}_m \leftarrow \frac{\boldsymbol{\mu}_m}{\|\boldsymbol{\mu}_m\|_2}.
    \label{eq:centroid}
\end{equation}
The representative entities of the scene are then selected by ranking words in $C_m$ according to their cosine similarity with $\boldsymbol{\mu}_m$.

In this way, we construct a compact set of scene-level clusters that group semantically coherent nouns, providing structured context for downstream tasks.

\paragraph{LLM-Driven Composition and Dataset Generation.}
To generate the final complex $\text{Prompt}$ (\text{Combo}), each background chunk is paired with a foreground entity from two distinct domains: 100 celebrities from $\text{VGGFace}$ and 200 non-person entities from the $\text{T2I-CompBench}$ dataset. 
Background chunks are obtained by segmenting each scene cluster into fixed-size slices of $K \in \{5,6\}$ nouns while preserving the original ordering within each cluster.

\paragraph{Round-Robin Pairing Strategy.}
Let $\mathcal{B}=\{B_\ell\}_{\ell=1}^{L}$ be the sequence of background noun chunks, where
$B_\ell = (w_{\ell,1},\dots,w_{\ell,K})$, and let 
$\mathcal{F}=\{f_1,\dots,f_{|\mathcal{F}|}\}$ denote the shuffled list of foreground entities.
Background chunks act as the primary index, and for each $\ell$ we assign
\begin{equation}
f_{j(\ell)} = f_{(\ell \bmod |\mathcal{F}|)+1},
\end{equation}
which ensures that every $B_\ell$ is matched with a foreground entity in a deterministic cyclic fashion.
Each paired tuple $(B_\ell, f_{j(\ell)})$ is then sent to a \textbf{Synthesis Agent} (GPT-4o) configured to produce \textbf{exactly two connected sentences}: the first sentence integrates all background nouns into a coherent scene description, and the second foreground sentence presents the entity $f_{j(\ell)}$ prominently.\\
\textbf{System Prompt for Synthesis Agent}\\
{\footnotesize
You write image scene descriptions. Input: (a) a list of nouns for BACKGROUND,
and (b) one OBJECT noun for FOREGROUND. Task: Compose all elements into ONE
coherent image, craft a coherent, natural-sounding visual description using
exactly two connected sentences. Create a clear and natural description of the
BACKGROUND by integrating all provided nouns; if some nouns are hard to combine,
group them into brief, meaningful phrases. You may add a few contextually
appropriate nouns to the BACKGROUND so it contains more than THREE distinct,
relevant nouns that enhance realism and clarity. Describe the OBJECT prominently
in the FOREGROUND so it stands out from the background. Output exactly TWO
sentences, concise, natural, and visual. No extra notes or meta text.
}\\
\textbf{Example} An illustrative instance from our dataset demonstrates how a background noun chunk and a foreground entity are synthesised into a concrete two-sentence description.
Given:\\
\textbf{User Prompt}\\
{\footnotesize
\textbf{BG nouns}:\emph{light, Lights, candlelight, sun, flame} \\
\textbf{Obj nouns}: \emph{Ken Davitian} \\
}
\textbf{the Synthesis Agent outputs}:
\emph{``The room is softly illuminated by a mixture of natural sunlight
streaming through the window and the gentle flicker of candlelight,
casting warm glows and subtle shadows on the walls.
In the foreground, Ken Davitian sits comfortably in an armchair,
the flames from the nearby candles reflecting in his eyes as he smiles warmly.''}

\subsection{Metrics}
\label{metrics}
We employ a comprehensive, multi-tiered evaluation suite to rigorously assess model performance across diverse dimensions. To evaluate Compositional Integrity and Semantic Decoupling, we utilize decomposed $\text{CLIP}$ scores that quantify alignment under simultaneous, fine-grained constraints. 
This structural fidelity is further validated by the Noun Recall score, which measures the successful grounding of both foreground and background entities, denoted by $\mathcal{N}_{\text{obj}}$ and $\mathcal{N}_{\text{bg}}$. We define
\begin{equation}
\dy{\mathcal{N}} = \mathcal{N}_{\text{obj}} \cup \mathcal{N}_{\text{bg}}.
\end{equation}
For each noun $n \in N$, its maximum detection confidence is given by
\begin{equation}
s(n) = \max_q p(q \mid n),
\end{equation}
where $q$ indexes predicted bounding boxes from an open-vocabulary detector~\cite{minderer2022simple}. Given a confidence threshold $\tau = 0.25$, the Noun Recall is defined as
\begin{equation}
\mathrm{NounRecall}
= \frac{1}{|\dy{\mathcal{N}}|} \sum_{n \in N} \mathbf{1}\big(s(n) \ge \tau\big).
\end{equation}
To capture human-aligned visual quality, we adopt HPSv3~\cite{ma2025hpsv3widespectrumhumanpreference}, while BLIP Diversity~\cite{li2022blip} measures intra-class variation. The latter is computed as the mean pairwise cosine distance between text embeddings of generated captions, reflecting the model's capacity for descriptive richness. Finally, we evaluate identity preservation using ID-Sim based on ArcFace~\cite{deng2019arcface}. To account for practical failure modes, images where no valid face is detected are assigned a penalty score of $-1$, ensuring a strict assessment of identity consistency.
\section{Prompt Decomposition and Validation}
\label{appendix:decomposition}

This appendix provides the full prompts, an illustrative example, and the additional technical details supporting the \emph{Decomposer Agent} introduced in Sec.~\ref{Long Prompt Decomposition}. Our goal is to transform a freeform two-sentence description into (i) a background sentence $\mathcal{P}_{\text{bg}}$, (ii) a main-object sentence $\mathcal{P}_{\text{obj}}$, (iii) an object noun set $\mathcal{N}_{\text{obj}}$, and (iv) a background noun set $\mathcal{N}_{\text{bg}}$.
The Decomposer Agent operates under strict verbatim constraints: all nouns must appear in the input composition, adjectives and verbs are removed, and no hallucinated entities are permitted. These structured components are subsequently used to initialise cross-attention masks for controllable diffusion.\\
\textbf{System Prompt For Decomposer Agent}
You are a meticulous decomposer. Input is a two-sentence composition for an image. Task:\\ 
1) BACKGROUND\_SENTENCE: one simple sentence ONLY about background, environment, time, objects and atmosphere. \\
2) OBJECT\_SENTENCE: one simple sentence ONLY about the main object and MUST include the target NAME exactly as given if present. \\
3) OBJECT\_NOUNS: 5-10 nouns copied verbatim from the Composition referring to the main object. \\
4) BACKGROUND\_NOUNS: 5-10 nouns copied verbatim from the Composition referring to locations/time/scenery/background objects/ambience. \\
Constraints: use ONLY words present in the Composition for noun lists; lowercase; no verbs/adjectives; no words not present; no extra keys. Return a compact JSON object with keys: background\_sentence, object\_sentence, object\_nouns, background\_nouns.

\subsection{Example of Prompt Decomposition}

We provide one full example to illustrate how freeform text is transformed into structured representations.

\paragraph{Input Description.}
\begin{quote}
``The room is softly illuminated by a mixture of natural sunlight streaming through the window and the gentle flicker of candlelight, casting warm glows and subtle shadows on the walls. In the foreground, Ken Davitian sits comfortably in an armchair, the flames from the nearby candles reflecting in his eyes as he smiles warmly.'' 
\end{quote}

\paragraph{Decomposition Output.}
\begin{itemize}
    \item \textbf{Background Sentence:}
    
    ``The room is softly illuminated by a mixture of natural sunlight streaming through the window and the gentle flicker of candlelight, casting warm glows and subtle shadows on the walls.''
    \item \textbf{Object Sentence:}
    
    ``Ken Davitian sits comfortably in an armchair, the flames from the nearby candles reflecting in his eyes as he smiles warmly.''
    \item \textbf{Background Nouns:}
    
    ["room", "sunlight", "window", "candlelight", "glows", "shadows", "walls", "flames", "candles"]
    \item \textbf{Object Nouns:}
    
    ["ken davitian", "foreground", "armchair", "eyes", "smile"]
\end{itemize}

These noun sets serve as semantic anchors for constructing cross-attention masks during controllable generation.

% --------------------------------------------------
\subsection{Candidate Generation and Scoring-Based Validation}
\label{appendix:validation}

The Decomposer Agent may generate imperfect candidates due to inherent model stochasticity. To ensure structural correctness and linguistic fidelity, we apply a two-stage validation procedure summarised below.

\paragraph{Multiple-Candidate Sampling}
For a given composition $\mathcal{C}$, the agent is queried under two temperature settings $t\in\{0.4, 0.9\}$:
\begin{equation}
\{\hat{D}^{(t)}\} = \text{Decomposer}(\mathcal{C}, t).
\end{equation}
Low temperature encourages syntactic correctness, while high temperature allows more complete noun enumeration.

If both attempts fail validation, two fallback temperatures $t\in\{0.2, 1.0\}$ are used.

\paragraph{Verbatim Noun Consistency}
Each returned JSON object is validated against the input composition.  
Let $\mathcal{C}_{\text{lower}}$ denote the lowercase composition string.  
For a candidate decomposition $D$, we compute:
\begin{align}
\mathcal{N}^{\text{valid}}_{\text{obj}} &=
\{ n \in \mathcal{N}_{\text{obj}} \mid n \subset \mathcal{C}_{\text{lower}} \}, \\
\mathcal{N}^{\text{valid}}_{\text{bg}} &=
\{ n \in \mathcal{N}_{\text{bg}} \mid n \subset \mathcal{C}_{\text{lower}} \},
\end{align}
followed by uniqueness filtering while preserving order.

Candidates containing fewer than two valid object nouns or two valid background nouns receive a negative score and are rejected.

\paragraph{Scoring Function}
For each candidate $D$, we compute:
\begin{equation}
s(D) = 
\begin{cases}
-1 & \text{if invalid},\\[2pt]
1 + |\mathcal{N}^{\text{valid}}_{\text{obj}}| + |\mathcal{N}^{\text{valid}}_{\text{bg}}|
    & \text{otherwise.}
\end{cases}
\end{equation}
The additive constant ensures that valid sentence separation yields a strictly higher score than noun-only matches.

The final decomposition is:
\begin{equation}
D^\star = \arg\max_D ~ s(D).
\end{equation}

\paragraph{Deterministic Fallback}
In the unlikely event that all four candidates fail:
\[
D^\star = \{\text{empty sentences and empty noun sets}\}.
\]
This ensures the pipeline never breaks.

% --------------------------------------------------
\subsection{JSON Structure and Format Constraints}

The Decomposer Agent is required to return JSON with exactly the following keys:

\texttt{background\_sentence},\quad
\texttt{object\_sentence},\quad
\\
\texttt{object\_nouns},\quad
\texttt{background\_nouns}.

No additional keys or commentary are allowed.  
The noun lists must contain only:
\begin{itemize}
    \item lowercase tokens,
    \item verbatim substrings of the input composition,
    \item nouns referring specifically to either background or the main object.
\end{itemize}

This strict format ensures compatibility with our downstream parsing and mask-generation modules.

\subsection{LLM-based agent robustness}
\deng{All 300 prompts were decomposed successfully with valid JSON and satisfied constraints; safeguards were implemented but never triggered. We verified robustness across both ChatGPT and Gemini, indicating that prompt formalization yields stable behaviour. }

\section{Hyperparameters}
\label{sec:hyper}

Unless otherwise specified, all models follow the same Dynamic Spatial Guidance and Dynamic Pruning procedure described in Sec.~\ref{Dynamic Pruning Mechanism}. 
We summarize only the hyperparameters that affect behavior or ablation results.

\paragraph{Backbones and precision.}
We use three backbones: \textbf{Stable Diffusion 2.1 base}, \textbf{Stable Diffusion XL base 1.0}, and \textbf{FLUX.1~dev}. 
Mixed precision is enabled when supported (bfloat16/float16 on GPU, otherwise float32), and all VAEs use standard tiling/slicing for memory efficiency.

\paragraph{Sampling and multi-path exploration.}
For all backbones we run $T{=}30$ denoising steps with $K{=}10$ stochastic trajectories in parallel, and apply pruning every $I{=}5$ steps.  
Classifier-free guidance (CFG) scales and image resolutions are:
\begin{itemize}
    \item SD2.1: $512{\times}512$, $\mathrm{CFG}{=}7.5$;
    \item SDXL: $1024{\times}1024$, $\mathrm{CFG}{=}7.5$;
    \item FLUX: $1024{\times}1024$, $\mathrm{CFG}{=}3.0$.
\end{itemize}
Each trajectory maintains an independent scheduler (\texttt{DDIMScheduler} for SD2.1/SDXL and \texttt{FlowMatchEulerDiscreteScheduler} for FLUX).

\paragraph{Latent-space background projection.}
Background anchoring follows Eq.~\eqref{eq:projection} with a piecewise-linear decay in Eq.~\eqref{eq:lambda_schedule}.  
For timestep index $i$ and $r=i/(T{-}1)$, we use
\begin{equation}
\lambda(i) = \Lambda \Bigl(1-\frac{r}{r_{\max}}\Bigr)\mathbb{1}\{r<r_{\max}\},
\end{equation}
where $(\Lambda,r_{\max})$ are backbone-specific:
\begin{itemize}
    \item SD2.1: $\Lambda{=}0.10$, $r_{\max}{=}0.20$;
    \item SDXL: $\Lambda{=}0.10$, $r_{\max}{=}0.20$;
    \item FLUX: $\Lambda{=}0.25$, $r_{\max}{=}0.35$.
\end{itemize}
Projection is always applied only on background locations using the soft mask $\widehat{M}_{\mathrm{bg}}$.

\paragraph{DSG mask construction and token grouping.}
For all backbones, background and object masks are obtained from cross-attention probabilities by (i) averaging over heads and selected layers, (ii) aggregating over the corresponding token sets $\mathcal{T}_{\mathrm{bg}}$ and $\mathcal{T}_{\mathrm{obj}}$ (Eqs.~\eqref{eq:avg_heads}–\eqref{eq:mask_fusion}), (iii) upsampling to latent resolution, and (iv) thresholding and smoothing.  
We use a single threshold $\tau$ and one iteration of $3{\times}3$ average-pooling smoothing:
\begin{itemize}
    \item SD2.1 / SDXL: $\tau{=}0.50$;
    \item FLUX: $\tau{=}0.40$.
\end{itemize}
If only one token set is non-empty, the other mask is taken as its complement.  
Object and background token sets are obtained by regex matching over decoded tokens using word lists $\mathcal{N}_{\text{obj}}$ and $\mathcal{N}_{\text{bg}}$ from the decomposition agent.

\paragraph{Internal token gating.}
We apply multiplicative token gating inside cross-attention with maximum gain $\gamma_{\max}{=}2.0$.  
During the \emph{background} stage we use gains $(+1.0$ for background tokens, $-0.5$ for object tokens$)$, and swap the roles in the \emph{object} stage.  
The effective gain decays with the relative step $r$ and is nonzero only in the early background window and the mid object window.

\paragraph{CLIP scoring and previews.}
For all backbones, CLIP scoring in Eq.~\eqref{eq:clip_score} uses ViT-B/32 on $256{\times}256$ preview images decoded from intermediate latents.  
The score averages cosine similarities between the image embedding and embeddings of the prompt clauses $\mathcal{P}_{\mathrm{clauses}}$.

\paragraph{Online pruning.}
Dynamic pruning uses the composite score in Eq.~\eqref{eq:score},
\begin{equation}
S = w_{\mathrm{tail}}\cdot \mathrm{norm}(-\mathrm{TailH})
  + w_{\mathrm{clip}}\cdot \mathrm{norm}(\mathrm{CLIP}),
\end{equation}
with identical weights $w_{\mathrm{tail}}{=}1.0$ and $w_{\mathrm{clip}}{=}2.0$ across all backbones.  
Tail entropy $\mathrm{TailH}$ is computed over the top $20\%$ of entropy values in both pixel-to-token and token-to-pixel directions.  
Pruning removes paths whose $S$ falls below a backbone-specific quantile threshold $q$, while always retaining at least one active trajectory:
\begin{itemize}
    \item SD2.1: $q{=}0.85$ (more conservative due to weaker backbone);
    \item SDXL / FLUX: $q{=}0.35$.
\end{itemize}

\paragraph{Sensitivity ranges.}
Empirically useful ranges are $\mathrm{CFG}\in[2.5,3.0]$ for FLUX and $\mathrm{CFG}\in[6.0,8.0]$ for SD2.1/SDXL, $\Lambda\in[0.15,0.35]$, $\tau\in[0.35,0.55]$, $\gamma_{\max}\in[1.0,2.0]$, pruning interval $I\in[5,8]$. \deng{For the number of paths, we evaluate $K\in[2,22]$ (Fig.~\ref{fig:clip_i_k}) and observe a consistent improvement in CLIP-I as $K$ increases. However, manual inspection of the final samples suggests that overly large $K$ can lead to noticeably more homogeneous generations (with occasional qualitative gains), while incurring substantial additional compute. We therefore suggest $K{=}10$ by default as a practical trade-off between efficiency and strong output quality.} 
Excessively large CFG or gating gain tends to oversharpen attention and reduce diversity, while too large $\Lambda$ may suppress fine foreground details.

\section{Additional Results}
\label{sec:appendix_additional_results}

\subsection{Quantitative Ablation Study}
\label{sec:appendix_ablation_quant}

In the main paper we qualitatively ablate three key components of our framework: Background Projection, DSG Gating, and Dynamic Pruning on SDXL and FLUX backbones. 
Here we provide a quantitative counterpart. 
Table~\ref{tab:ablation_study_quantitative} reports global mean scores of all evaluation metrics used in the main paper.

For clarity, we use the following configurations:
\begin{itemize}
    \item \textbf{w/o BGproj}: disables the latent-space Background Projection in Eq.~\eqref{eq:projection} by setting $\lambda(t)\equiv 0$ for all timesteps.
    \item \textbf{w/o Dsg}: removes DSG Gating and uses a uniform mask $\widehat{M}_{\mathrm{bg}}\equiv 1$, forcing global correction.
    \item \textbf{w/o Prune}: disables Dynamic Pruning entirely, i.e., all $K$ trajectories are kept until the final decoding.
    \item \textbf{w/o clipScore}: keeps the pruning schedule but removes the CLIP term by setting $w_{\mathrm{clip}}=0$ in Eq.~\eqref{eq:score}; pruning is driven solely by attention Tail Entropy.
    \item \textbf{w/o Entropy}: symmetrically removes the Tail Entropy term by setting $w_{\mathrm{tail}}=0$, so pruning is guided only by CLIP alignment.
    \item \textbf{OursSDXL} / \textbf{OursFLUX}: full models with all components enabled as described in Sec.~\ref{Dynamic Pruning Mechanism}.
\end{itemize}

\begin{table}[ht]
    \centering
    \caption{Quantitative ablation of core components on SDXL and FLUX backbones.}
    % \resizebox{\textwidth}{!}{
    \begin{tabular}{lccccc}
    \toprule
    \textbf{Method} & \textbf{CLIP\_I} & \textbf{BLIP} & \textbf{NounRecall} & \textbf{HPSv3} \\
    \midrule
    \multicolumn{5}{l}{\textbf{SDXL backbone}} \\
    w/o BGproj    & 0.3475 & 0.4827 & 0.5687 &  8.3981 \\
    w/o Dsg       & 0.2780 & 0.3456 & 0.4903 &  1.0348 \\
    w/o Prune     & 0.3487 & 0.4858 & 0.5718 &  8.4099 \\
    w/o clipScore & 0.3456 & 0.4806 & 0.5713 &  8.3277 \\
    w/o Entropy   & 0.3521 & 0.5066 & 0.5751 &  7.9871 \\
    OursSDXL      & \textbf{0.3538} & \textbf{0.5091} & \textbf{0.5884} & \textbf{8.8063} \\
    \midrule
    \multicolumn{5}{l}{\textbf{FLUX backbone}} \\
    w/o BGproj    & 0.3517 & 0.5052 & 0.6074 & 10.9766 \\
    w/o Dsg       & 0.1747 & 0.1742 & 0.2211 &  0.7332 \\
    w/o Prune     & 0.3494 & 0.4915 & 0.6139 & 11.3271 \\
    w/o clipScore & 0.3452 & 0.4697 & 0.6001 & 11.9355 \\
    w/o Entropy   & 0.3425 & 0.4997 & 0.5687 & 10.9748 \\
    OursFLUX      & \textbf{0.3520} & \textbf{0.5152} & \textbf{0.6389} & \textbf{11.9849} \\
    \bottomrule
    \end{tabular}
    % }
    \label{tab:ablation_study_quantitative}
\end{table}
Overall, removing any component degrades performance.
For SDXL, eliminating DSG Gating (\textbf{w/o Dsg}) causes the largest drop across all metrics, confirming that spatially localized projection is essential for safe latent editing.

\begin{figure}[t]
    \centering
    \includegraphics[width=\linewidth]{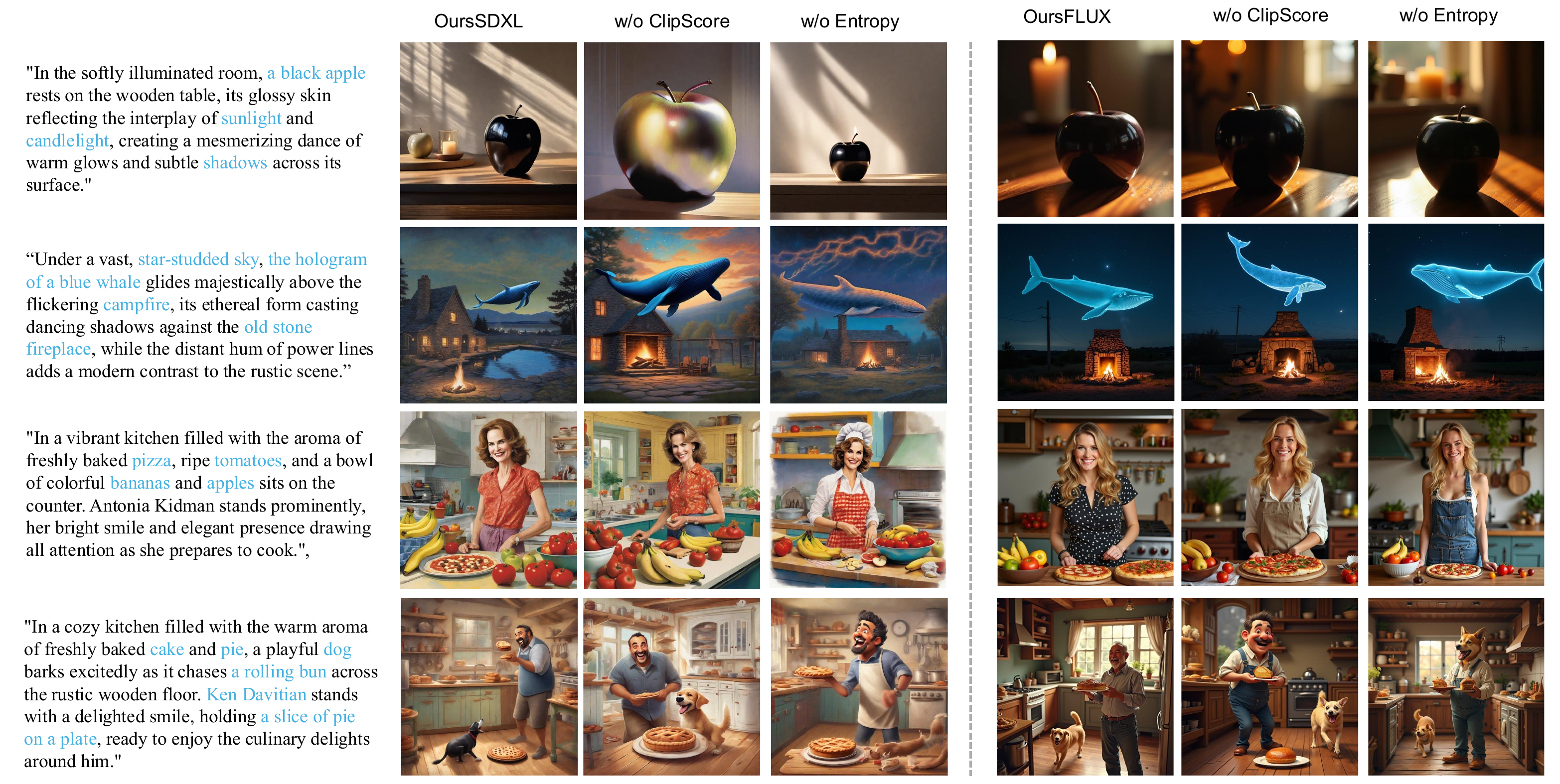}
    \vspace{-3mm}
    \caption{
    Qualitative ablation of the Dynamic Pruning signals on SDXL (left block) and FLUX (right block).
    For each prompt, we show the full model (\textbf{Ours}), the variant without CLIP-based scoring (\textbf{w/o clipScore}, i.e., $w_{\mathrm{clip}}{=}0$), and the variant without Tail Entropy (\textbf{w/o Entropy}, i.e., $w_{\mathrm{tail}}{=}0$).
    Removing the CLIP term often preserves coarse layout but fails to satisfy fine-grained textual details (e.g., missing secondary objects or incorrect background attributes), while removing the entropy term leads to visually noisy or structurally inconsistent regions due to unstable attention patterns. 
    When both signals are used jointly, the selected trajectories exhibit sharper foregrounds and more faithful, text-aligned backgrounds.
    }
    \label{fig:appendix_pruning_qual}
\end{figure}
\FloatBarrier
\begin{center}
    \includegraphics[width=0.7\linewidth]{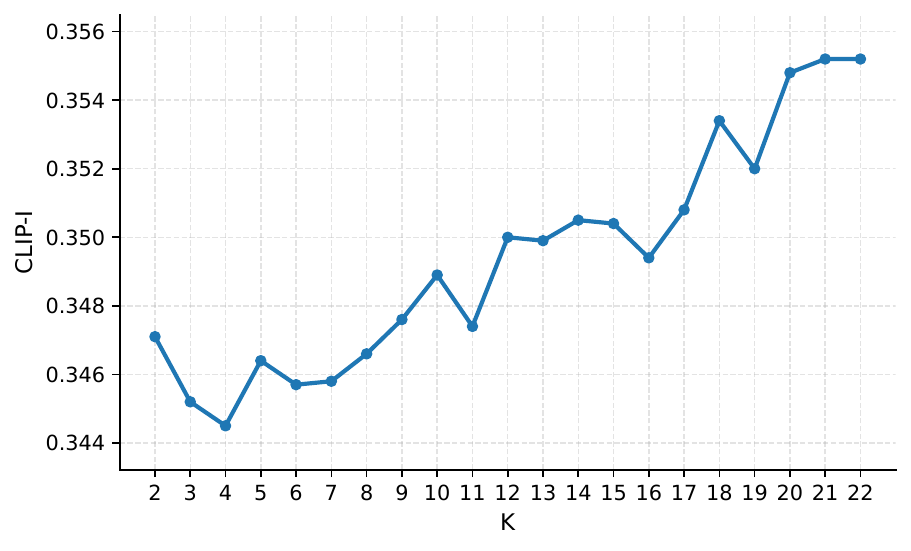}
    \captionsetup{type=figure}
    \vspace{-2mm}
    \caption{CLIP-I score as a function of the number of paths $K$.}
    \label{fig:clip_i_k}
\end{center}
Disabling Background Projection (\textbf{w/o BGproj}) or Dynamic Pruning (\textbf{w/o Prune}) also reduces CLIP and BLIP scores compared to the full model.
For the pruning-related variants, \textbf{w/o clipScore} and \textbf{w/o Entropy} both underperform \textbf{OursSDXL}, indicating that attention Tail Entropy and CLIP alignment provide complementary signals.

A similar trend holds for FLUX: \textbf{w/o Dsg} almost collapses all scores, \textbf{w/o BGproj} and \textbf{w/o Prune} show consistent degradation, and removing either CLIP or entropy weakens semantic grounding (NounRecall, BLIP) as well as human preference (HPSv3). 
Across both backbones, the full models (\textbf{OursSDXL} and \textbf{OursFLUX}) achieve the best or near-best results on all metrics.

\subsection{Qualitative Ablation of Dynamic Pruning Signals}
To complement the quantitative analysis, we further visualize the effect of the two scoring terms used in Dynamic Pruning (Eq.~\ref{eq:score}).
Figure~\ref{fig:appendix_pruning_qual} presents side-by-side comparisons of the full model (\textbf{Ours}) with \textbf{w/o clipScore} and \textbf{w/o Entropy} on both SDXL and FLUX backbones.
\end{document}